\documentclass[a4paper,twoside]{article}

\usepackage{epsfig}
\usepackage{subfigure}
\usepackage{calc}
\usepackage{amssymb}
\usepackage{amstext}
\usepackage{amsmath}
\usepackage{amsthm}
\usepackage{multicol}
\usepackage{pslatex}
\usepackage{apalike}
\usepackage{siunitx}

\usepackage{amsmath}
\usepackage{amsfonts}
\usepackage{graphicx}
\graphicspath{{images/}}
\usepackage{subfigure}
\usepackage{url,multirow,array,color}
\usepackage[procnumbered,ruled]{algorithm2e}

\usepackage{SCITEPRESS}     % Please add other packages that you may need BEFORE the SCITEPRESS.sty package.

\begin{document}

\title{From Depth Data to Head Pose Estimation: a Siamese approach}

% \author{\authorname{Omitted for blind review}}
%\author{\authorname{Anonymized related submission - Submitted at VISAPP}}

\author{\authorname{Marco Venturelli, Guido Borghi, Roberto Vezzani, Rita Cucchiara}%
\affiliation{University of Modena and Reggio Emilia, DIEF \\ Via Vivarelli 10, Modena, Italy}%
\email{\{marco.venturelli, guido.borghi, roberto.vezzani, rita.cucchiara\}@unimore.it}}

\keywords{Head Pose Estimation, Deep Learning, Depth Maps, Automotive}

\abstract{
The correct estimation of the head pose is a problem of the great importance for many applications. For instance, it is an enabling technology in automotive for driver attention monitoring.
In this paper, we tackle the pose estimation problem through a deep learning network working in regression manner. Traditional methods usually rely on visual facial features, such as facial landmarks or nose tip position. In contrast, we exploit a Convolutional Neural Network (CNN) to perform head pose estimation directly from depth data. We exploit a Siamese architecture and we propose a novel loss function to improve the learning of the regression network layer. The system has been tested on two public datasets, \textit{Biwi Kinect Head Pose} and \textit{ICT-3DHP database}. The reported results demonstrate the improvement in accuracy with respect to current state-of-the-art approaches and the real time capabilities of the overall framework. 
}

\onecolumn \maketitle \normalsize \vfill

\section{\uppercase{Introduction}}
\label{sec:introduction}

\noindent Head pose estimation provides a rich source of information that can be used in several fields of computer vision, like attention and behavior analysis, saliency prediction and so on. In this work, we focus in particular on the automotive field: several works in literature show that head pose estimation is one of the key elements for driver behavior and attention monitoring analysis. Moreover, the recent introduction of semi-autonomous and autonomous driving vehicles and their coexistence with traditional cars is going to increase the already high interest on driver attention studies. In this cases, human drivers have to take driving algorithms under controls, also for legal related issues \cite{Rahman4021}.

Driver's hypo-vigilance is one of the most principal cause of road crashes \cite{alioua2016}. As reported by the official US government website\footnote{http://www.distraction.gov/index.html}, distracting driving is responsible for 20-30\% of road deaths: it is reported that about 18\% of injury crashes were caused by distraction, more than 3000 people were killed in 2011 in a crash involving a distracted driver, and distraction is responsible for 11\% of fatal crashes of drivers under the age of twenty. Distraction during driving activity is defined by \textit{National Safety Administration} (NHTSA) as "\textit{an activity that could divert a person's attention away from the primary task of driving}".
\cite{CrayeK15} defines three classes of driving distractions: 1) \textit{manual distraction}: driver's hands are not on the wheel; examples of this kind of activity are incorrect use of infotainment system (radio, GPS navigation device and others) or text messaging. 2) \textit{visual distraction}: driver's eyes are not looking at the road, but, for example, at the smart-phone screen or a newspaper. 3) \textit{cognitive distraction}: driver's attention is not focused on driving activity; this could occur due to torpor, stress, and bad physical conditions in general or, for example, if talking with passengers.
Smartphone abuse during driving activity leads to all of the three distraction categories mentioned above; in fact, that is one of the most important cause of fatal driving distraction, with about 18\% of fatal driver accidents in North America, as reported by NHTSA.\\
Several works have been proposed for in-car safety and they can be divided by the type of signal used \cite{alioua2016}. \\
1) \textit{Physiological signals}: special sensors as electroencephalography (EEG), electrocardiography (ECG) or electromyography (EMG) are places inside the cockpit to acquire signals from driver's body; this kind of solution is very intrusive and a body-sensor contact is strictly required; \\
2) \textit{Vehicle signals}: vehicle parameters like velocity changes, steering wheel motion, acquired from car bus, can reveal abnormal driver actions;\\ 
3) \textit{Physical signals}: image processing techniques are exploited to investigate driver vigilance through facial features, eye state, head pose or mouth state; these methods are non-intrusive, thus image are acquired from inside cockpit cameras. \\
Taking into account the above exposed elements, some characteristics can be elected as crucial for a reliable and implementable head pose estimation framework, also related to the placement and the choice of the most suitable sensing device: 
\begin{itemize}
\item \textbf{Light invariance}: the framework should be reliable on each weather condition that could dramatically changes the type of illumination  inside the car (shining sun and clouds, in addition to sunrises, sunsets, nights etc.). Depth cameras are proven to be less prone to fail in these conditions than classical RGB or stereo sensors;
\item \textbf{Non invasive}: it is fundamental that acquisition devices do not impede driver's movements during driving activity; in this regard, recently many car industries have placed sensors inside steering wheel or seats to passively monitor driver's physiological conditions;
\item \textbf{Direct estimation}: the presence of severe occlusions or the high variability of driver's body pose could make facial feature detection extremely challenging and prone to failure; besides, no initialization phase is welcome.
\item \textbf{Real time performances}: in automotive context an attention monitoring system is useful only if can immediately detect anomalies in driver's behavior; 
\item \textbf{Small size}: acquisition device  has to been integrated inside cockpit, often in a particular position (like next rear-view mirror): recently, the release of several low cost,  accurate and small sized 3D sensors open new scenarios.
\end{itemize}
In this work, we aim at exploiting a deep architecture to perform in real time head pose regression, directly from single-frame depth data. In particular, we use a \textit{Siamese} network to improve our training phase by learning more discriminative features, and optimize our regression layer network loss function.

\section{\uppercase{Related work}}
\begin{figure}[t!]
\centering
\subfigure[]{\includegraphics[width=0.25\columnwidth]{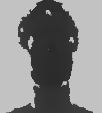}} 
\subfigure[]{\includegraphics[width=0.25\columnwidth]{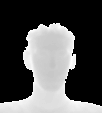}} 
\subfigure[]{\includegraphics[width=0.25\columnwidth]{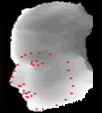}}
\subfigure[]{\includegraphics[width=0.25\columnwidth]{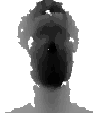}} 
\subfigure[]{\includegraphics[width=0.25\columnwidth]{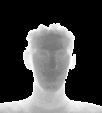}} 
\subfigure[]{\includegraphics[width=0.25\columnwidth]{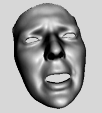}} 
\caption{Examples of depth images taken by different acquisition devices. (a) is acquired by \textit{Microsoft Kinect} based on structured-light technology (BIWI dataset \cite{fanelli2011}). (b) is obtained thanks to \textit{Microsoft Kinect One}, a time-of-flight 3D scanner; (d)-(e) are the correspondent images, after contrast stretching elaboration to enhance facial clues. Images (c)-(f) come from synthetic dataset \cite{eth_biwi_00760,baltruvsaitis2012}.} 
\label{fig:sampledepth} 
\end{figure}

\noindent \cite{Trivedi2009} shows that head pose estimation is the goal of several works in the literature. Current approaches can be divided depending on the type of data they rely on, RGB images (2D information), depth maps data (3D information), or both. In general, methods for head pose estimation relying solely on RGB images are sensitive to illumination, partial occlusions and lack of features \cite{fanelli2011}, while depth-based approaches are lacking of texture and color information. \\
Several works in the literature proposed to use Convolutional Neural Networks with depth data, but especially in skeleton body pose estimation \cite{crabbe2015skeleton} or action recognition tasks \cite{ji20133d}. These works reveal how techniques like background subtraction, depth maps normalization and data augmentation could influence deep architectures performance. Recently, \cite{DoumanoglouBKK16} exploits Siamese Networks to perform object pose estimation, applying a novel loss function that can boost the performance of a regression network layer. Other works, like \cite{hoffer2015deep,sun2014deep}, exploit a Siamese approach in deep architecture to improve network learning capabilities and to perform human body joint identification.\\
%DEPTH 
Several works rely only on depth data. As Figure \ref{fig:sampledepth} shows, the global quality of depth images strictly depends by the technology of the acquisition device. In \cite{papazov2015} shapes of 3D surfaces are encoded in a novel triangular surface patch descriptor to map an input depth with the most similar ones that were computed from synthetic head models, during a precedent a training phase.
\cite{kondori2011} exploits a least-square minimization of the difference between the rate prediction and the measured rate of change of input depth.
Usually, depth data are characterized by low quality. Starting from this assumption, in \cite{malassiotis2005} is proposed a method designed to work on low quality depth data to perform head localization and pose estimation; this method relies on an accurate nose  localization.
In \cite{fanelli2011} a real time framework based on Random Regression Forests is proposed, to perform head pose estimation directly from depth images, without exploiting any facial features.
In \cite{padeleris2012} Particle Swarm Optimization is used to tackle the head pose estimation,  treated as an optimization problem. This method requires an initial frame to construct the reference head pose from depth data; limited real time performance are obtained thanks to a GPU.\\
\cite{breitenstein2008} tackles the problem of large head pose variations, partial occlusions and facial expressions from depth images: several methods present in the literature have poor performance with these factors. In the work of Bretenstein et al. the main issue is that the nose must be always visible, due to this method uses geometric features to generate nose candidates which suggest head position hypothesis. The alignment error computation is demanded to a dedicated GPU in order to work in real time.
\begin{figure*}[t!]
    \centering
    \includegraphics[width=0.85\linewidth]{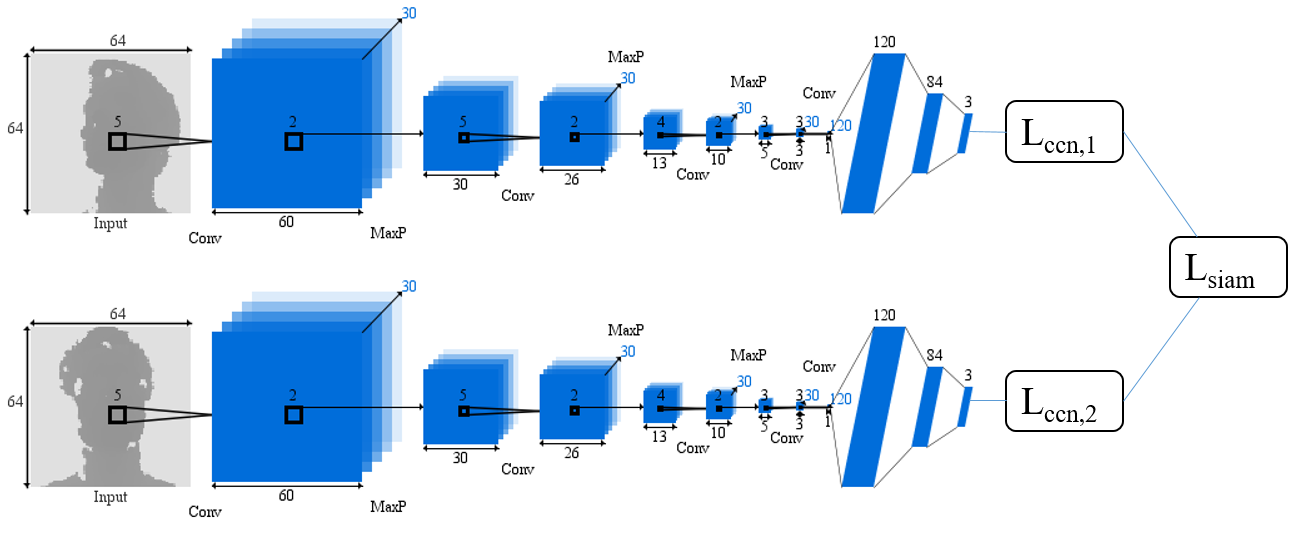}
    \caption{The Siamese architecture proposed for training phase.}
    \label{fig:siamese}
\end{figure*}
%RGB
\cite{chen2016} achieves results very close to state-of-art results, even if in this work the problem of head pose estimation is token on extremely low resolution RGB images.
HOG features and a Gaussian locally-linear mapping model are used in \cite{drouard2015}. These models are learned using training data, to map the face descriptor onto the space of head poses and to predict angles of head rotation.\\
A Convolutional Neural Network (CNN) is used in \cite{ahn2014} to perform head pose estimation from RGB images. This work shows that a CNN properly works even in challenging light conditions. The network inputs are RGB images acquired from a monocular camera: this work is one of the first attempt to use deep learning techniques in head pose estimation problem. This architecture is exploited in a data regression manner to learn the mapping function between visual appearance and three dimensional head estimation angles. Despite the use of deep learning techniques, system working real time with the aid of a GPU. A CNN trained on synthetic RGB images is used also in \cite{liu3d2016}. Recently, the use of synthetic dataset is increasing to support deep learning approaches that basically require huge amount of data.
%DEPTH + RGB 
A large part of works relies on both 2D and 3D data. In \cite{Seemann2004} a neural network is used to combine depth information, acquired by a stereo camera, and skin color histograms derived from RGB images. The user face has to be detected in frontal pose at the beginning of framework pipeline to initialize the color skin histograms.
In \cite{baltruvsaitis2012} a 3D constrained local method for robust facial feature tracking under varying poses is proposed. It is based on the integration both depth and intensity information.\\
\cite{bleiweiss2010} used time-of-flight depth data to perform a real time head pose estimation, combined with color information. The computation work is demanded to a dedicated GPU.
\cite{yang2012} elaborated HOG features both on 2D and 3D data: a Multi Layer Perceptron is then used for feature classification. Also the method presented in \cite{saeed2015} is based on RGB and depth HOG, but a linear SVM is used for classification task.
Ghiass et al. \cite{ghiass2015} performed pose estimation by fitting a 3D morphable model which included pose parameter, starting both from RGB and depth data. This method relies on face detector of Viola and Jones \cite{viola2004}.

\section{\uppercase{Head Pose Estimation}}
\noindent The described approach aims at estimating pitch, roll and yaw angles of the head/face with respect to the camera reference frame. A depth image is provided as input and a Siamese CNN is used to build an additional loss function which improves the strength of the training phase. Head detection and localization are supposed to be available. No additional information such as facial landmarks, nose tip position, skin color and so on are taken into account, differently from other methods like \cite{Seemann2004,malassiotis2005,breitenstein2008}. The network prediction is given in terms of Euler angles, even if the task is challenging due to problems such periodicity \cite{yi2015} and the non-continuous nature of Euler angles \cite{kendall2015posenet}.

\subsection{Head acquisition}
\noindent First of all, face images are cropped using a dynamic window.  
Given the center $x_c, y_c$ of the face, each image is cropped at a rectangular box centered in $x_c, y_c$, with width and height computed as: 
\begin{equation}
w, h = \frac{f_{x,y} \cdot R}{Z},
\label{eq:wh}
\end{equation}
\noindent where $f_{x,y}$ are the horizontal and vertical focal lengths (in pixels) of the acquisition device, $R$ is the width of a generic face (300 mm in our experiments) and $Z$ is the distance between the acquisition device and the user obtained from the depth image. The output is an image which contains a partially centered face and some part of background. Then, the cropped images are resized to 64x64 pixels. 
Input image values are normalized to set their mean and the variance to 0 and 1, respectively. This normalization is also required by the specific activation function of the network layers.

%%%%%%%%%%%%%%%%%%%%%%%%%%%%%%%%%%%%%%%%%%%%%%%%%%%%%%%%%%%%%%
%% TABELLA EXPERIMENTS MESSA QUI PER STILE 
%%%%%%%%%%%%%%%%%%%%%%%%%%%%%%%%%%%%%%%%%%%%%%%%%%%%%%%%%%%%%%
\begin{table*}[h]
\caption{Results on \textit{Biwi Dataset}: pitch, roll and yaw are reported in Euler angles.}
\centering
\small
%\begin{tabular}{|p{0.8cm}|p{1.2cm}|c|c|c|p{0.9cm}|}
\begin{tabular}{|c|c|c|c|c|}
\hline
\textbf{Method} & \textbf{Data} &\textbf{Pitch} & \textbf{Roll}  & \textbf{Yaw}       \\ \hline
\cite{saeed2015}              & RGB+RGB-D 			& 5.0 $\pm$ 5.8			& 4.3 $\pm$ 4.6				& 3.9 $\pm$ 4.2					  \\ \hline
\cite{fanelli2011}            & RGB-D 					& 8.5 $\pm$ 9.9			& 7.9 $\pm$ 8.3			& 8.9 $\pm$ 13.0	  \\ \hline
\cite{yang2012}				  & RGB+RGB-D				& 9.1 $\pm$ 7.4			& 7.4 $\pm$ 4.9			& 8.9 $\pm$ 8.2		 \\ \hline
\cite{baltruvsaitis2012}      & RGB+RGB-D				& 5.1 					& 11.2					& 6.29		\\ \hline
\cite{papazov2015}      	  & RGB-D				    & 3.0 $\pm$ 9.6 		& 2.5 $\pm$ 7.4			& 3.8 $\pm$ 16.0 	\\ \hline
Our            				  & RGB-D					& 2.8 $\pm$ 3.2			& 2.3 $\pm$ 2.9			& 3.6 $\pm$ 4.1		\\ \hline
\textbf{Our+Siamese }         		  & RGB-D					& \textbf{2.3 $\pm$ 2.7}			& \textbf{2.1 $\pm$ 2.2}			& \textbf{2.8 $\pm$ 3.3}		\\ \hline
\end{tabular}
\label{tab:resBiwiPose}
\end{table*}
%%%%%%%%%%%%%%%%%%%%%%%%%%%%%%%%%%%%%%%%%%%%%%%%%%%%%%%%%%%%%%

\subsection{Training phase}
\noindent The proposed architecture is depicted in Figure \ref{fig:siamese}. A Siamese architecture consists in two or more separate networks, that could be identical --- as in our case--- and are simultaneously trained. It is important to note that this Siamese architecture is used only during training phase, while a single network is used during the testing. Inspired by \cite{ahn2014}, each single neural network has a shallow deep architecture in order to obtain real time performance and good accuracy. Each network takes images of 64x64 pixels as input and it is composed of 5 convolutional layers. The first four layers have 30 filters each,  whereas the last one has 120 filters. Max-pooling is conducted only three times, due to the relative small size of input images. At the end of the network there are three fully connected layers, with 120, 84 and 3 neurons, respectively. The last 3 neurons correspond to the three angles (\textit{yaw}, \textit{pitch} and \textit{roll}) of the head. The last fully connected layer works in regression. The size of the convolution filters are 5x5, 4x4, 3x3, depending on the layer. The activation function is the hyperbolic tangent (\textit{tanh}): in this way, the network can map output $[-\infty, +\infty] \rightarrow [-1, +1]$, even if ReLU tends to train faster that other activation functions \cite{krizhevsky2012}. The network is able to output continuous instead of discrete values. We adopt the Stochastic Gradient Descent (SGD) as in \cite{krizhevsky2012} to solve the back-propagation.\\ 
Each single neural network has a L2 loss:
\begin{equation}
L_{cnn} = \sum_i^n \lVert y_i - f(x_i) \rVert ^ 2 _2,
\label{eq:loss1}
\end{equation}
\noindent where $y_i$ is the ground truth information (expressed in roll, pitch and yaw Euler angles) and $f(x_i)$ is the network prediction.\\
Siamese network takes in input pair of images: considering a dataset with about $N$ frames, a huge number $\binom{N}{2}$ of possible pairs can be used. Only pairs with at least 30 degrees of difference between all head angles are selected.\\
Exploiting Siamese architecture, an additional loss function based on both network outputs can be defined. This loss combines each of the two regression losses and it is the L2 distance between the prediction difference and the ground truth difference: 
\begin{equation}
\begin{array}{rcl}
L_{siam} & =& \sum_i^n \lVert d_{cnn}(x_i) - d_{gt}(x_i) \rVert ^ 2 _2 \\ 
d_{cnn}(x_i) &=& f_1(x) - f_2(x) \\ 
d_{gt}(x_i) &=& y_1 - y_2 
\end{array},
\label{eq:losses}
\end{equation}
\noindent where $d_{cnn}(x_i)_k$ is the difference between the outputs $f_i(x)$of the two single networks and $d_{gt}(x_i)$ the difference between the ground truth values of the pair.\\
The final loss is a combination of the losses function of the 2 single networks $L_{cnn, 1}, L_{cnn, 2}$ and the loss of the Siamese match $L_{siam}$:
\begin{equation}
L=L_{cnn,1}+L_{cnn,2}+L_{siam}
\label{eq:lossfinal}
\end{equation}

\noindent Each single network has been trained with a batch size of 64, a decay value of $5^{-4}$, a momentum value of $9^{-1}$ and a learning rate set to $10^{-1}$, decreased up to $10^{-3}$ in the final epochs \cite{krizhevsky2012}. Ground truth angles are normalized to $[-1, +1]$.\\
We performed data augmentation to increment the size of training input images and to avoid over fitting. Additional patches are randomly cropped from each corner of the input images and from the head center; besides, patches are also extracted by cropping input images starting from the bottom, upper, left and right and adding Gaussian noise. Other additional input samples are created thanks to the pair input system: different pairs are different inputs for the Siamese architecture, and so also for each single network. Besides, data augmentation conducted in this manner produces samples with occlusion, and thus our method could be reliable against head occlusions. \\
%RV: non è bello dire qui "secondo noi".... In our opinion the main benefits of our Siamese approach during training phase are the novel loss function that boost learning performance of the regression layer and the considerable number of possible input pairs that, in addition with data augmentation, increase the dataset size and its semantic asset. \\
% As mentioned above, Our Siamese architecture is used only during training phase.

\section{\uppercase{Experimental results}}
\label{sec:experiments}

\noindent Experimental results of the proposed approach are given using two public Kinect datasets for head pose estimation, namely \textit{Biwi Kinect Head Pose Database} and \textit{ICT-3DHP database}. Both of them contains RGB and depth data. To check the reliability of proposed method we performed a cross-dataset validation, training the network on the first dataset and testing on the second one. The evaluation metric is based on the \textit{Mean Average Error} (MAE) between the absolute difference in angle between network predictions and ground truth.

\subsection{Biwi Kinect Head Pose Database}
Introduced in \cite{fanelli2013}, it is explicitly designed for head pose estimation  from depth data. About 15000 upper body images of 20 people (14 males and 6 females; 4 people were recorded twice) are present. The head rotation spans about $\pm \ang{75}$ for yaw, $\pm \ang{60}$ for pitch and $\pm \ang{50}$ for roll. Both RGB and depth images are acquired sitting in front a stationary \textit{Microsoft Kinect}, with a resolution of 640x480. Besides ground truth pose angles, calibration matrix and head center - the position of the nose tip - are given. Depth images are characterized by visual artifacts, like holes (invalid values in depth map). In the original work \cite{fanelli2011}, the total number of samples used for training and testing and the subject selection is not clear. We use sequences 11 and 12 to test our network, which correspond to not repeated subjects. Some papers use own method to collect results (e.g. \cite{ahn2014}), so their results are not reported and analyzed.

\subsection{ICT-3DHP Database}
\textit{ICT-3DHP Dataset} \cite{baltruvsaitis2012} is a head pose dataset, collected using \textit{Microsoft Kinect} sensor. It contains about 14000 frames (both intensity and depth), divided into 10 sequences. The resolution is 640x480. The ground truth is annotated using a \textit{Polhemus Fastrack} flock of birds tracker, that require a showy white cap, well visible in both RGB and RGB-D frames. This dataset is not oriented for deep learning, because of its small size and the presence of few subjects. 

\subsection{Quantitative evaluation}
The performance of the proposed head pose estimation are compared with a baseline system. To this aim, we trained a  single network with the structure of one Siamese component. Input data and data augmentation are the same on both cases. In addition, the results are also compared with other state-of-the-art techniques. As above mentioned, the training has been done on \textit{Biwi dataset} (2 subjects used for test), while the testing phases also exploited the \textit{ICT-3DHP dataset}. 

\begin{table*}[tbh]
\caption{Results on \textit{ICT-3DHP Dataset}: pitch, roll and yaw are reported in Euler angles.}
\centering
\small
%\begin{tabular}{|p{0.8cm}|p{1.2cm}|c|c|c|p{0.9cm}|}
\begin{tabular}{|c|c|c|c|c|}
\hline
\textbf{Method} & \textbf{Data} &\textbf{Pitch} & \textbf{Roll}  & \textbf{Yaw}       \\ \hline
\cite{saeed2015}              & RGB+RGB-D 				& 4.9 $\pm$ 5.3			& \textbf{4.4} $\pm$ 4.6			& \textbf{5.1} $\pm$ \textbf{5.4}		  \\ \hline
\cite{fanelli2011}            & RGB-D 					& 5.9 $\pm$ 6.3			& - 					& 6.3 $\pm$ 6.9	  \\ \hline
\cite{baltruvsaitis2012}      & RGB+RGB-D				& 7.06 					& 10.48					& 6.90				\\ \hline
Our            				  & RGB-D					& 5.5 $\pm$ 6.5			& 4.9 $\pm$ 5.0			& 10.8 $\pm$ 11.0		\\ \hline
\textbf{Our+Siamese}            		  & RGB-D					& \textbf{4.5} $\pm$ \textbf{4.6}			& \textbf{4.4} $\pm$ \textbf{4.5}			& 9.8 $\pm$ 10.1		\\ \hline
\end{tabular}
\label{tab:res3dhp}
\end{table*}

Table \ref{tab:resBiwiPose} reports the experimental results obtained on \textit{Biwi Kinect Head Pose Dataset}. The evaluation protocol is the same proposed in \cite{fanelli2011}.
Results reported in Table \ref{tab:resBiwiPose} show that our method overcomes other state-of-the-art techniques, even those working on both RGB and depth data. 

\begin{figure*}[h]
    \centering
    \includegraphics[width=1\linewidth]{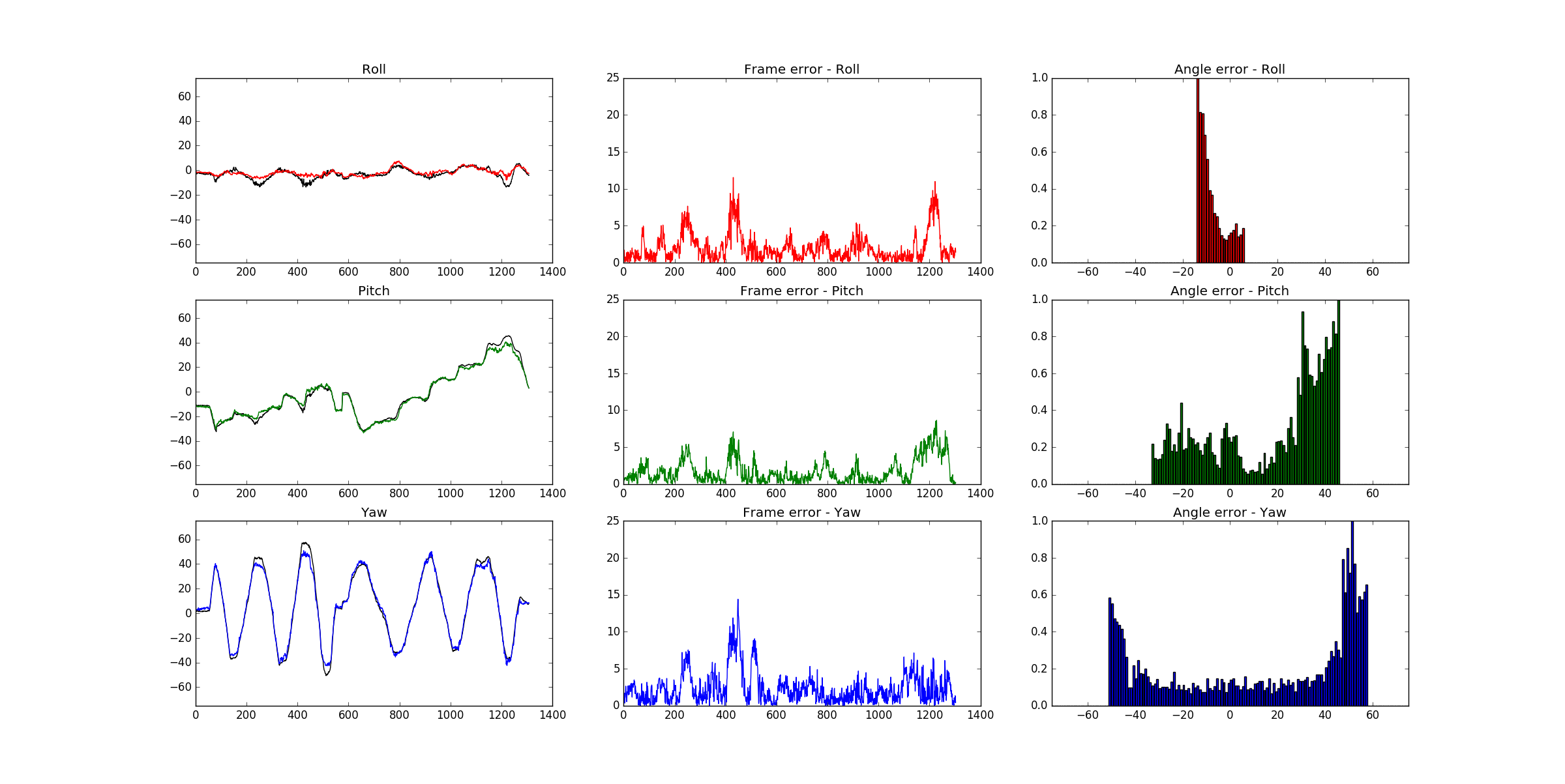}
    \caption{Experimental results for \textit{Biwi dataset}: ground truth is black. The second column reports the angle error per frame, while the last column reports histograms that highlight the errors at specific angles.}
    \label{fig:resultsBiwiGraphs}
\end{figure*}

\begin{figure*}[h!]
    \centering
    \includegraphics[width=1\linewidth]{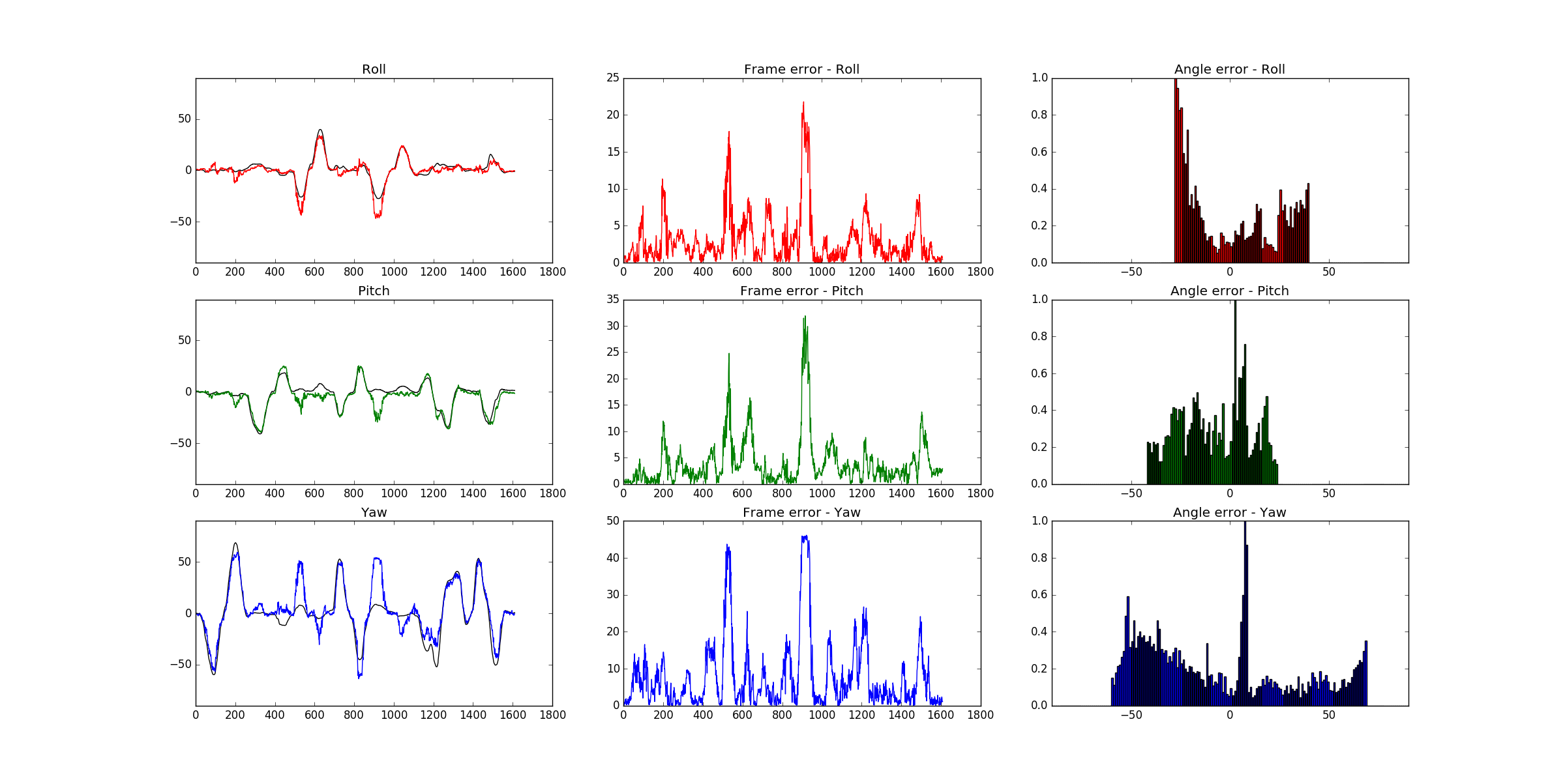}
    \caption{Experimental results for \textit{ICT-3DHP dataset}: see Figure \ref{fig:resultsBiwiGraphs} for explanation. }
    \label{fig:resultsICTGraphs}
\end{figure*}

\begin{figure*}[h!]
    \centering
    \includegraphics[width=1.0\linewidth]{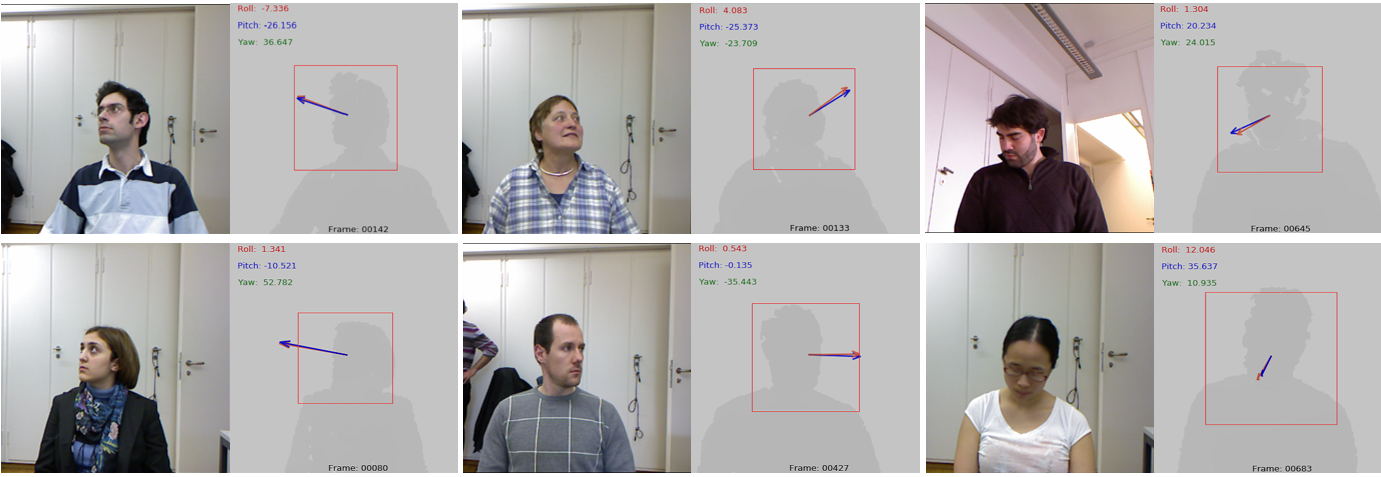}
    \caption{The first and the third columns show RGB frames, the second and the fourth the correspondent depth map frame with a red rectangle that reveals the crop for the face extraction (see Section 3.1). The blue arrow is ground truth, while the red one is our prediction. Numerical angle prediction is reported on the top. Images taken from \textit{Biwi Dataset}.}
    \label{fig:resultsBiwiImages}
\end{figure*}

Table \ref{tab:res3dhp} reports the results on \textit{ICT-3DHP Dataset}; the values related to \cite{fanelli2011} were taken from \cite{crabbe2015skeleton}. On this dataset, the dynamic face crop algorithm is degraded due to an imprecise head center location provided in the available ground truth. The authors published the position of the device exploited to capture the head angle instead of the head itself. Thus, part of the head center locations are inaccurate. To highlight this problem, we report that \cite{baltruvsaitis2012} had a substantial improvement of using GAVAM \cite{morency2008}, an adaptive key frame based differential tracker, over all other trackers. Their method in this case reports an absolute error of 2.9 for yaw, 3.14 for pitch and 3.17 for roll. 
Finally, we highlight the benefit of Siamese training phase. In fact, the proposed approach perform better than the single network as well as the other competitors, even those which rely on both RGB and RGB-D data. The prediction of roll angles is accurate, even if in the second dataset there is a lack of training data images with roll angles.\\
Figure \ref{fig:resultsBiwiGraphs} and Figure \ref{fig:resultsICTGraphs} report angle frame per error and errors at specific angles for both dataset. \\
Figure \ref{fig:resultsBiwiImages} shows an example of working framework for head pose estimation in real time: head center is taken thanks to ground truth data; the face is cropped from raw depth map (in the center image, the blue rectangle) and in the right frame yaw, pitch and roll angles are shown.
Total time of processing on a CPU (\textit{Core i7-4790} 3.60GHz) is 11.8 s and on a GPU (\textit{NVidia Quadro k2200}) is 0.146 s, computed on 250 frames from \textit{Biwi}.

% \begin{table}[h]
% \centering
% \small
% %\begin{tabular}{|p{0.8cm}|p{1.2cm}|c|c|c|p{0.9cm}|}
% \begin{tabular}{|c|c|c|}
% \hline
% \textbf{frames} 			  & \textbf{CPU} 			&\textbf{GPU}    					 \\ \hline
% 250             & 2.94 s 				& 0.036 s						\\ \hline
% 500				  & 5.91 s				& 0.073 s					  	\\ \hline
% 1000      	  & 11.8				    & 0.146 s				     \\ \hline
% \end{tabular}
% \caption{Processing time for our head pose estimation framework for 250, 500 and 1000 frames. The GPU used is a \textit{Nvidia Quadro k2200}}
% \label{tab:time}
% \end{table}

\begin{table}[b]
\caption{Performance evaluation (\textit{fps})}
\centering
\small
%\begin{tabular}{|p{0.8cm}|p{1.2cm}|c|c|c|p{0.9cm}|}
\begin{tabular}{|c|c|c|}
\hline
\textbf{Method} 			  & \textbf{Time} 				& \textbf{GPU}   		\\ \hline
\cite{fanelli2011}            & 40 ms/frame 				& x						\\ \hline
\cite{papazov2015}			  & 76 ms/frame					& 					  	\\ \hline
\cite{yang2012} 	     	  & 100 ms/frame				&  				     	\\ \hline
\textbf{Our}				  & \textbf{10 ms/frame}			& x						\\ \hline
\end{tabular}
\label{tab:time_method}
\end{table}

\section{Conclusion}
\noindent We present a innovative method to directly extract head angles from depth images in real time, exploiting  a deep learning approach. Our technique aim to deal with two main issue of deep architectures in general, and CNNs in particular: the difficulty to solve regression problems and the traditional heavy computational load that compromise real time performance for deep architectures.
Our approach is based on Convolutional Neural Network with shallow deep architecture, to preserve time performance, and is designed to resolve a regression task.\\
There is rich possibility for extensions thanks to the flexibility of our approach: in future work we plan to integrate temporal coherence and stabilization in the deep learning architecture, maintaining real time performance, incorporate RGB or infrared data to investigate the possibility to have  a light invariant approach even in particular conditions (e.g. automotive context). Head localization through deep approach could be studied in order to develop a complete framework that can detect, localize and estimate head pose inside a cockpit.
%bruttino finire così: Besides studies about how occlusions can deprecate our method are being conducted.

%\vfill
\bibliographystyle{apalike}
{\small
\bibliography{2015VISAPP_scene}}

%\vfill
\end{document}